\documentclass[letterpaper, 10 pt, conference]{ieeeconf}  % Comment this line out if you need a4paper

\IEEEoverridecommandlockouts                              % This command is only needed if 
\usepackage{amssymb}  % assumes amsmath package installed
\usepackage{graphicx}
\usepackage{amsmath}
\usepackage{caption}
\usepackage{subcaption}
\usepackage{cleveref}
\usepackage{gensymb}
\usepackage{pgfgantt}
\usepackage{tabularx}
\usepackage{cellspace}
\usepackage{lipsum}
\usepackage{tikz}
\usepackage{pgfplots}
\usepackage[backend=biber]{biblatex} % Use biber as the backend
\usepackage{svg}
\usepackage{adjustbox}
\usepackage{multirow}

\title{\LARGE \bf Online Diffusion-Based 3D Occupancy Prediction at the Frontier \\ with Probabilistic Map Reconciliation}

\author{Alec Reed, Lorin Achey, Brendan Crowe, Bradley Hayes, Christoffer Heckman% <-this % stops a space
\thanks{This work was partially supported by the National Science Foundation award \#2339328.}% <-this % stops a space
\thanks{All authors are affiliated with Department of Computer Science, University of Colorado Boulder, USA. E-mail addresses for authors are:
        {\tt\small FirstName.LastName@colorado.edu}}%
}

\begin{document}

\maketitle
\thispagestyle{empty}
\pagestyle{empty}

%%%%%%%%%%%%%%%%%%%%%%%%%%%%%%%%%%%%%%%%%%%%%%%%%%%%%%%%%%%%%%%%%%%%%%%%%%%%%%%%
\begin{abstract}
Autonomous navigation and exploration in unmapped environments remains a significant challenge in robotics due to the difficulty robots face in making commonsense inference of unobserved geometries. Recent advancements have demonstrated that generative modeling techniques, particularly diffusion models, can enable systems to infer these geometries from partial observation. In this work, we present implementation details and results for real-time, online occupancy prediction using a modified diffusion model. By removing attention-based visual conditioning and visual feature extraction components, we achieve a 73$\%$ reduction in runtime with minimal accuracy reduction. These modifications enable occupancy prediction across the entire map, rather than being limited to the area around the robot where camera data can be collected. We introduce a probabilistic update method for merging predicted occupancy data into running occupancy maps, resulting in a 71$\%$ improvement in predicting occupancy at map frontiers compared to previous methods. Finally, we release our code and a ROS node for on-robot operation $\langle$upon publication$\rangle$ at github.com/arpg/sceneSense$\_$ws.
\end{abstract}

%%%%%%%%%%%%%%%%%%%%%%%%%%%%%%%%%%%%%%%%%%%%%%%%%%%%%%%%%%%%%%%%%%%%%%%%%%%%%%%%
\section{INTRODUCTION}
In general robots are limited to evaluating and making decisions over space that has been directly observed, either during the present deployment or a prior one. For deployments to environments where prior information does not exist, the autonomous system relies only on what it can observe at present. These deployments are particularly challenging for autonomous navigation as perception sensors have limited fields of view, and are often occluded by obstacles in the environment. Data products, such as 2D or 3D geometric maps that are generated for these environments, can have holes where the sensor could not observe, particularly at runtime when the system is exploring. While there are existing methods for filling these gaps, most focus on filling LIDAR shadows \cite{lidarShadowFilling} or gaps in the map \cite{xu2019depthcompletionsparselidar, cheng2020s3cnetsparsesemanticscene} where the geometry around the target area has been observed. To further enhance robotic decision making, we not only need to fill holes and gaps in the map, but also extend map geometries beyond what can be directly measured.
\begin{figure}[ht!]
\vspace{10pt}
    \centering
    \includegraphics[width=\linewidth]{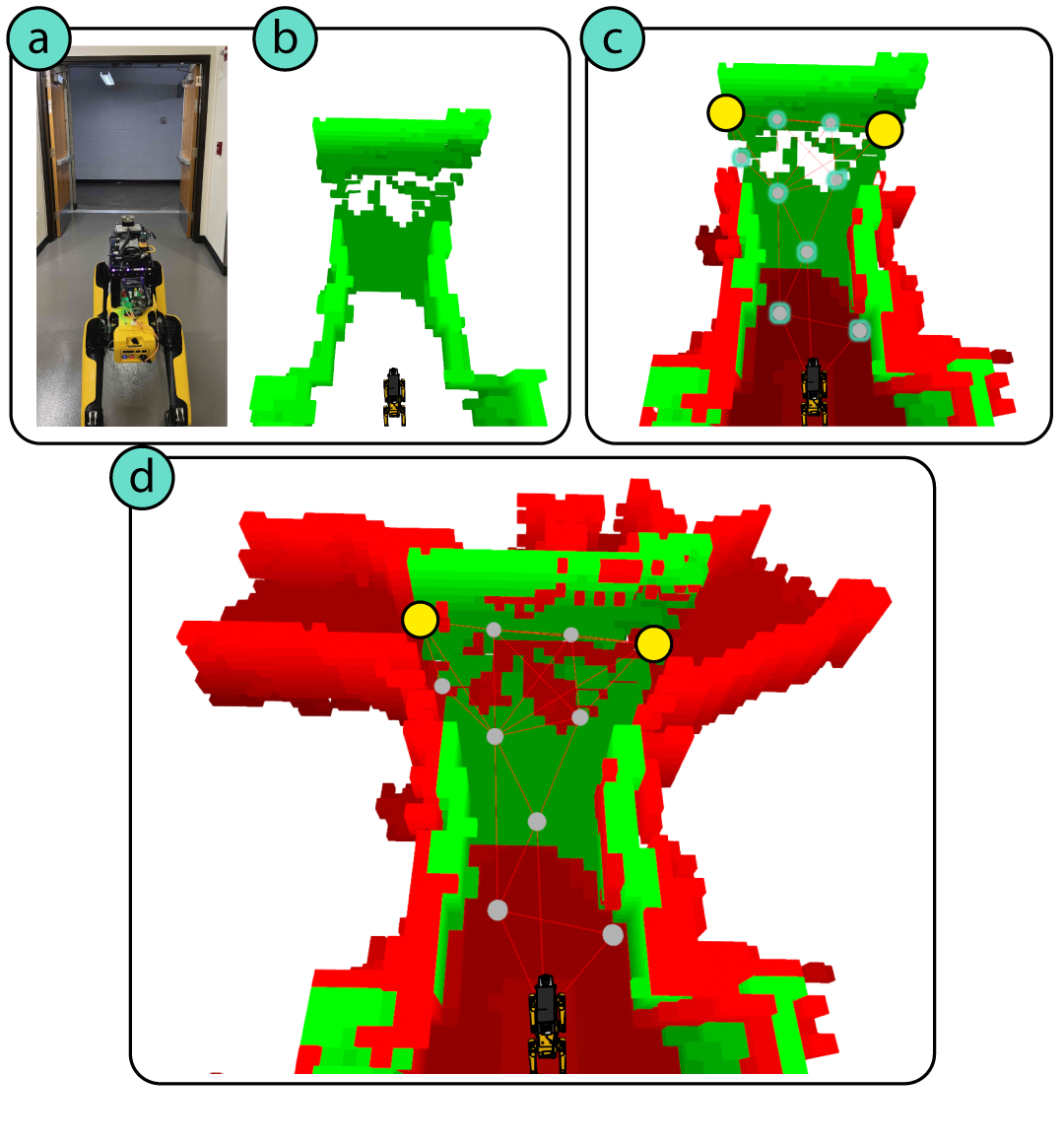}
    \caption{\textbf{Onboard Occupancy Prediction and Map Merging}: Green voxels represent observed occupancy and red voxels represent predicted occupancy. Gray graph points represent vertices and yellow graph points represent vertices identified as frontier points. \textbf{(a)} Spot platform is positioned in front of a t-intersection at startup as shown in the photo of the scene. \textbf{(b)} The map is populated with the observed 3D occupancy data from the LIDAR sensor. \textbf{(c)} Robot-centric (RC) occupancy prediction runs to predict occupancy data around the robot. Then a graph is built over the space to identify frontiers of interest for frontier-centric (FC) occupancy prediction. \textbf{(d)} Finally the diffusion model predicts the occupancy around the frontier points. These predicted maps are merged into the running map using our probabilistic map update rule.}
    \label{fig:main}
\end{figure}
Occupancy prediction is a method to fill and extend observed maps beyond direct measurements made by sensors. Recent works \cite{reed2024scenesense} have shown that occupancy prediction models can create realistic and likely predictions of what complete occupied space could look like around a robot. However it is not obvious how these methods would transfer from simulation to a real-world system. In this paper, we make key modifications to the diffusion-based SceneSense occupancy prediction model \cite{reed2024scenesense} to enable occupancy predictions at any point in the running map, as well as achieve decreased inference times for online occupancy prediction. Further, we define a probabilistic map update rule to merge the occupancy predictions with the running observed map. We implement a graph-based frontier evaluation method for identifying ideal areas for occupancy prediction and evaluate it with a real-world robotic system. 
The primary outcomes of the contributions discussed are:
\begin{enumerate}
    \item $73\%$ end-to-end run time reduction for SceneSense \cite{reed2024scenesense} occupancy prediction model. 
    \item Enabling occupancy predictions at range, anywhere in the map. 
    \item $75\%$ improvement in frontier occupancy map evaluation metrics when compared to the vision only map.
    \item $71\%$ improvement in frontier occupancy map evaluation metrics when compared to the one shot map merging method presented in the original SceneSense work \cite{reed2024scenesense}.
\end{enumerate}

\section{RELATED WORK}
\label{sec:related_works}

\subsection{Occupancy Prediction}
One solution to the challenge of autonomous navigation in occluded environments is to predict occupancy distributions. Deep learning (DL) approaches have shown promise, but existing methods struggle with scalability, generalization, and handling occluded areas. Wang et al. \cite{wang2021learning3doccupancyprediction} propose a DL approach to predict occupancy distribution which involves selectively removing data from the Matterport3D \cite{chang2017matterport3d} dataset during training for the model to learn occluded geometries. This biases the model to predict occupancy for these specific types of occlusions and does not scale well to large unseen sections of an environment. In \cite{Huang_2024_CVPR}, the authors present a self-supervised method for 3D occupancy prediction using video sequences, which transforms 2D images into 3D representations with deformable attention layers. While effective with nearby cameras, it struggles to predict occupancy beyond the camera's view due to signed distance field's (SDF) limitations in managing occluded geometry. 
% In \cite{reed2023looking}, unlabeled lidar point clouds are used to predict out-of-range or occluded road points, which are then combined with the original scan to create a terrain-extended point cloud. Our work expands upon this foundational approach, addressing key limitations identified in the original framework.
More recently, diffusion models were shown to successfully generate occupancy predictions behind occluded geometries in indoor environments using a single RGBD sensor mounted on a mobile robot platform \cite{reed2024scenesense}. Our research advances this method by introducing a novel, more efficient approach to occupancy prediction, demonstrated on real hardware.

\subsection{Scene Synthesis}
\label{sec:method}
\begin{figure}
\vspace{1pt}
    \centering
    \includegraphics[width=0.98\linewidth]{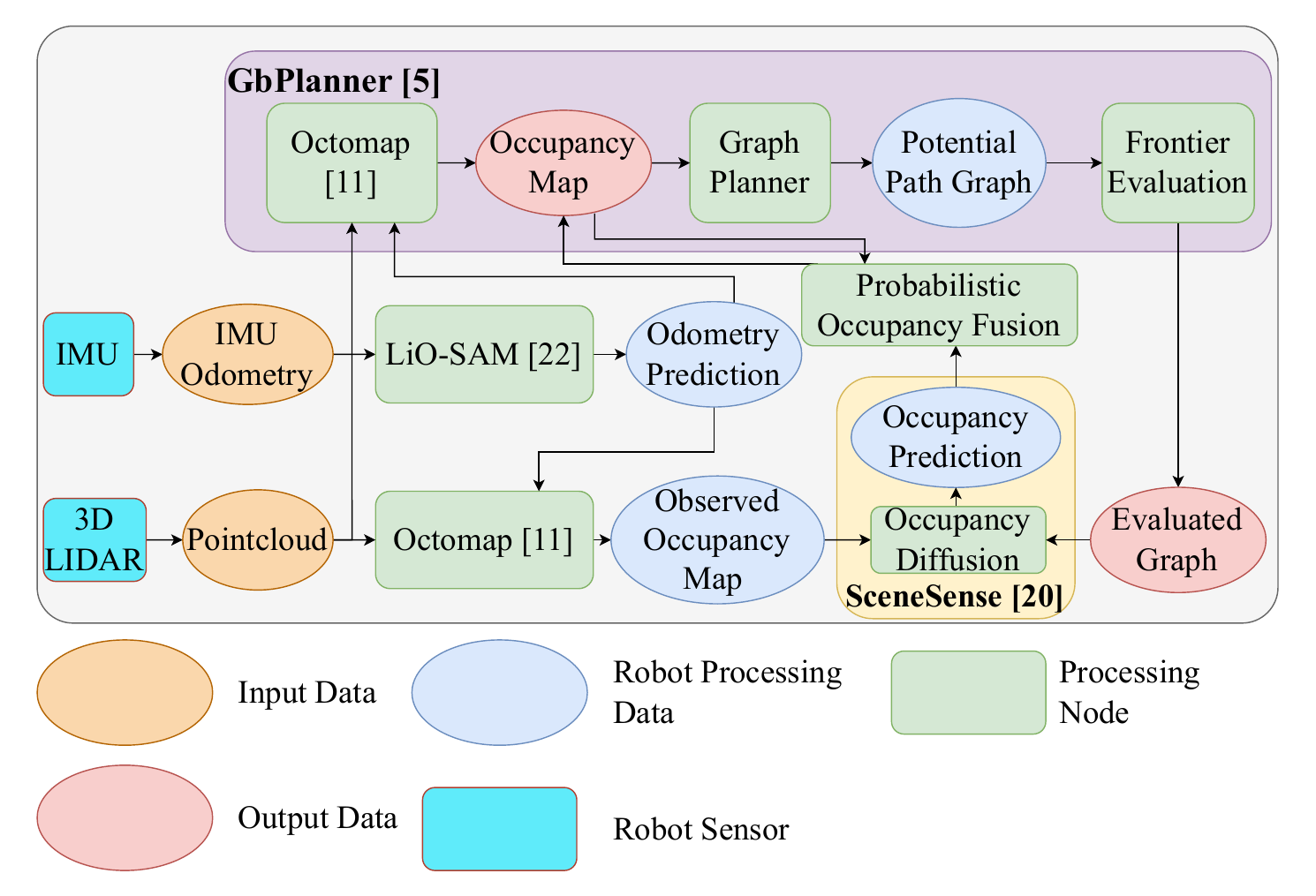}
    \caption{\textbf{System Block Diagram}:  Block diagram showing the system design for onboard SceneSense occupancy prediction. The system is comprised of an IMU and LIDAR sensor to generate odometry and occupancy maps. Once the occupancy map is built, a graph is constructed to evaluate frontier points for occupancy prediction. Local occupancy is then subselected around these points and sent to the SceneSense framework that provides occupancy predictions. These predictions are then merged with the running occupancy map using the probabilistic update rule.}
    \label{fig:robot_archetecture}
\end{figure}
Diffusion models \cite{ho2020denoising, sohl2015deep}, are a popular tool that has demonstrated impressive generative results across image \cite{rombach2022high}, video \cite{harvey2022flexible}, and natural language \cite{huang2022prodiff}. Based on these successes, diffusion models are being extended to 3D scene and shape generation. Recent work \cite{luo2021diffusionpointcloudgeneration} demonstrates the use of diffusion models for 3D point cloud generation for simple shapes and objects (e.g. tables, chairs). Kim et al. \cite{kim2023scenegendiffusionmodels} shows successful 3D shape generation from 2D content such as images, and Vahdat et al. \cite{vahdat2022lion} demonstrate similar 3D shape generation but using point cloud datasets rather than images. In LegoNet \cite{wei2023lego}, diffusion models are used to propose object rearrangements in a 3D scene. In DiffuScene \cite{tang2023diffuscene}, a denoising diffusion model is used with text conditioning to generate 3D indoor scenes from sets of unordered object attributes. Unlike these previous works which primarily focus on generating simple shapes, rearranging objects, or creating indoor scenes, our approach leverages diffusion models to fuse generated terrain with measurements from the local robot field of view, thereby bridging the gap between 3D scene generation and practical, situated robotics applications.
\begin{figure*}[t]
    \centering
    \includegraphics[width=\linewidth]{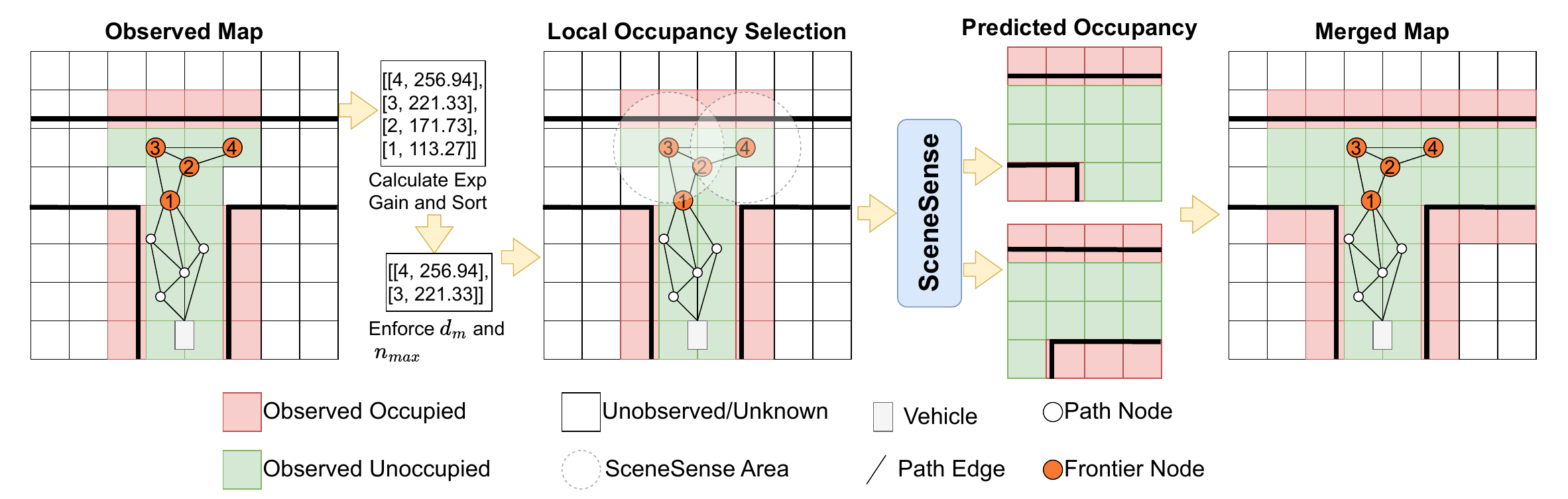}
    \caption{\textbf{Map Merging Example}: Process for generating and merging occupancy predictions with an observed map. A graph is generated and evaluated to identify frontier nodes. Then, the frontier nodes are sorted by exploration gain as defined in Eq. \ref{eq:GB_planner}, and $d_m$ (min node spacing) and $n_{max}$ (max frontier prediction nodes) are enforced on the frontier points set. For each frontier point identified for occupancy prediction, local occupancy is selected from the observed map and sent to SceneSense for occupancy prediction. Finally, the predicted maps are merged into the running occupancy map using Eq. \ref{eq:prob_map_merging}.}
    \label{fig:map_merge_cycle}
\end{figure*}
% \subsection{Generative AI in Robotics}
% The body of work on generative AI in robotics is steadily growing. Yuan et al. \cite{yuan2023hierarchicalgenmodeling} show that hierarchical generative modeling, inspired by human motor control, allows autonomous robots to perform complex tasks robustly. DALL-E-Bot \cite{kapelyukh2023DALLEBOT}, leverages diffusion models to generate images from text prompts, enabling robots to rearrange real objects accordingly. Diffusion models have also been applied to improve robot motion planning \cite{carvalho2023planningdiffusionmodels}. 

\section{METHODS}
\subsection{Problem Definition}
\textbf{Frontier Identification and Evaluation}. \ \ 
Let $\mathbb{M}$ be the current occupancy map, built via measurements from an onboard sensor $\mathbb{S}$ and odometer measurements $\mathbb{O}$.  The map consists of voxels $m$ that are categorized as $m \in \mathbb{M}_{free}$, $m \in \mathbb{M}_{occupied}$, or $m \in \mathbb{M}_{unknown}$, representing free, occupied and unknown space respectively. We seek to identify and evaluate frontiers in $\mathbb{M}$ that can enhance the robot's decision making for potential exploration. In general, ``interesting'' frontiers will maximize the number of unknown voxels available for occupancy prediction while considering common exploration metrics such a directionality, distance from target, and reachability \cite{biggie2023flexible, dang2020graph}.
 \vspace{5pt}\\
 % Given a map $\mathbb{M}$ find $n$ points $p \in \mathbb{R}^3$ where occupancy prediction can be performed to maximize \aar{not sure how to finish this}
\textbf{Dense Occupancy prediction}. \ \
Dense occupancy prediction predicts the occupancy from $[0,1]$ where 0 is unoccupied and 1 is occupied for every voxel $m$ in a target region $x \subset \mathbb{M}$. 

\subsection{Robotic System Architecture}
\label{sec:method_robo_sys_arch}
Our robotic system is constructed as a quadruped (Spot) as shown in Figure \ref{fig:main} as well as an off-board high performance computer to handle computationally expensive requests.  A block diagram of the system is shown in Figure \ref{fig:robot_archetecture}.
\vspace{5pt}\\
\textbf{Sensor Suite} \ \
The equipped sensor suite was designed with the purpose of providing 3D point cloud information and sufficient data for accurate localization. The primary sensor in the spot sensor suite is the Ouster OS1-64 LIDAR which provides 3D point clouds for mapping and localization. In addition a LORD Microstrain 3DM-GX5-15 IMU is used to measure the linear and angular acceleration of the Ouster, for use in the localization system. 
% In addition to the built on sensors the Boston Dynamics Spot also has 5 stereo pairs of depth cameras: two at the front, one at the rear, and one on each side of Spot.
\vspace{5pt}\\
\textbf{Localization}.\ \
Localization is required for Spot to perform volumetric mapping. We have implemented the popular LIDAR-based localization method LiO-SAM \cite{shan2020liosam} to provide localization at run-time. 
% \aar{should I cite the subt paper for this?} To improve localization performance both the IMU and lidar sensor were fasted to a 6061 aluminum sensor plate. This allowed for a high-precision relative transform between the sensors, reducing the necessity on extrinsic calibration. Further LIO-SAM requires aligned sensor timestamps and sensor publish rates to be constant. The LORD IMU however allows for large fluctuation in publish rate due to the USB transmission delays. TO reduce the sensitivity the IMU timesamps are adjusted when messages are not received within $15\%$ of the nominal rate. Additionally the lidar sensor uses PTP to synchronize with the onboard computer. These modifications allowed for consistent localization using LIO-SAM. 
\vspace{5pt}\\
\textbf{Occupancy Prediction}. \ \
We adopt the SceneSense occupancy prediction diffusion model \cite{reed2024scenesense} with modifications to enable novel functionality and performance improvements. Originally, SceneSense was designed as a conditional diffusion model where the conditioning was RGBD data from a camera/depth sensor on the robot. However, ablation studies of the model  show that including this RGBD conditioning has very little performance impact when \emph{occupancy inpainting} is enabled \cite{reed2024scenesense}. By removing the RGBD conditioning we enable two key characteristics for the model; anywhere occupancy prediction and increased inference speed.

Removing the RGBD conditioning data obviates the need to center occupancy predictions at the robot. By removing the need for image conditioning, occupancy can be predicted anywhere in the observed map, allowing for occupancy predictions at range. Secondly, we can replace the cross-attention based noise prediction model with the equivalent unconditional model. This reduces the number of trainable parameters for similar unconditioned performance. Further, this change also removes the need for a feature extraction backbone, saving additional computation time.
\vspace{5pt}\\
\textbf{Frontier Identification and Evaluation}. \ \ 
With these modifications to the occupancy prediction framework we can generate occupancy predictions at any point in the map. To identify interesting areas for prediction we adopt the graph-based exploration planner GBPlanner \cite{dang2020graph}. GBPlanner builds a graph where nodes are potential exploration points and edges are feasible paths to navigate from node to node. A ray casting algorithm is run at each node in the graph to quantify the number of observed, free, and unknown voxels in that node's field of view. After finding the shortest path to each node the \emph{Exploration Gain} can be calculated for each node in the graph as follows: 
\begin{multline}
\label{eq:GB_planner}
    \textbf{ExplorationGain}(\sigma_i) = e^{-\gamma_\mathcal{S}\mathcal{S}(\sigma_i,\sigma_{exp})}  \\
    \cdot \sum_{j = 1}^{m_i}\textbf{VolumetricGain}(v^i_j)e^{-\gamma_\mathcal{D}\mathcal{D}(v^i_1,v^i_j)},
\end{multline}
where $\mathcal{S}(\sigma_i, \sigma_{exp})$, $\mathcal{D}(v^i_1,v^i_j)$ are weight function with tunable factors $\gamma_\mathcal{S}, \gamma_\mathcal{D} > 0$ respectively. Furthermore $\mathcal{D}(v^i_1,v^i_j)$ is the cumulative Euclidean distance from a vertex $v^i_j$ to the root $v^i_1$ along a path $\gamma_i$.

Exploration gain is used to rank nodes at which occupancy prediction should run. Given a minimum node spacing $d_{m}$ and a maximum number of frontier prediction nodes $n_{max}$, SceneSense generates occupancy predictions at range, centered around the identified frontiers.
\vspace{5pt}\\
\textbf{Mapping}.
The probabilistic volumetric mapping method Octomap \cite{hornung13octomap} was selected as the mapping framework. Octomap was adopted for its log-odds update method for predicting occupied and unoccupied cells. This approach allows for elegant fusion of observed occupancy and predicted occupancy maps. Further discussion on map fusion is provided in Section \ref{sec:prob_map_merging}. 

\subsection{Probabilistic Map Merging}
\label{sec:prob_map_merging}
In previous work, predicted occupancy was merged into the running occupancy map in a ``fire and forget'' approach \cite{reed2024scenesense}. A given occupancy prediction was temporarily merged into the running occupancy map by setting the predicted occupied cells to 1. Then, when a new occupancy prediction was generated, the previous prediction would be removed from the running map and the new prediction would be merged in the same way. While this approach is effective in some applications, it limits the ability to accurately maintain a coherent and continuous understanding of the environment. To address these issues, we modify the probabilistic occupancy update formula presented for the Octomap mapping framework \cite{hornung13octomap}.

We define the probability that a voxel $m$ is occupied given an occupancy prediction $d_t$ or sensor reading $z_t$ as $P(m|d_t)$ or $P(m|z_t)$ respectively. The set of sensor estimates $z_{1:t}$ and diffusion estimates $d_{1:t}$ populate the joint set $\{z_{1:t}, d_{1:t}\}$ which we denote as  $j_{1:t}$. As discussed in \cite{reed2024scenesense}, SceneSense only operates on voxels $m$ that are not contained in the observed set $\mathcal{O}$, where $z_{t:t-1} = \varnothing$, and therefore $P(m|j_{1:t})$ will never require an update given $P(m|d_t)$ and $P(m|z_t)$ at the same time. As such we generate the piece-wise update rule for merging diffusion into the running occupancy map.

\begin{equation} \label{eq:prob_map_merging}
\begin{aligned}
P(m \mid j_{1:t})= \\
& \hspace{-60pt} \begin{cases} 
\left[ 1 + \frac{1-P(m \mid d_t)}{P(m \mid d_t)}\frac{1-P(m \mid j_{1:t-1})}{P(m \mid j_{1:t-1})} \frac{P(m)}{1-P(m)}\right]^{-1} & \text{if } m \notin \mathcal{O}  \\
\left[ 1 + \frac{1-P(m \mid z_t)}{P(m \mid z_t)}\frac{1-P(m \mid j_{1:t-1})}{P(m \mid j_{1:t-1})} \frac{P(m)}{1-P(m)}\right]^{-1} & \text{if } m \in \mathcal{O}.
\end{cases}
\end{aligned}
\end{equation}

In this framework $P(m|z_t)$ and $P(m|d_t)$ can be configured to different values prior to runtime. Generally  $P(m|d_t)$ given a predicted occupied cell is set lower than $P(m|z_t)$ given a sensed occupied cell, as we trust the sensor more than our generative model. By using this probabilistic approach to map merging, the final merged map benefits from prediction persistence as the system explores as well as increased map fidelity due to multi-prediction occupancy refinement.  
\begin{figure}[t]
\vspace{5pt}
    \centering
    \includegraphics[width=\linewidth]{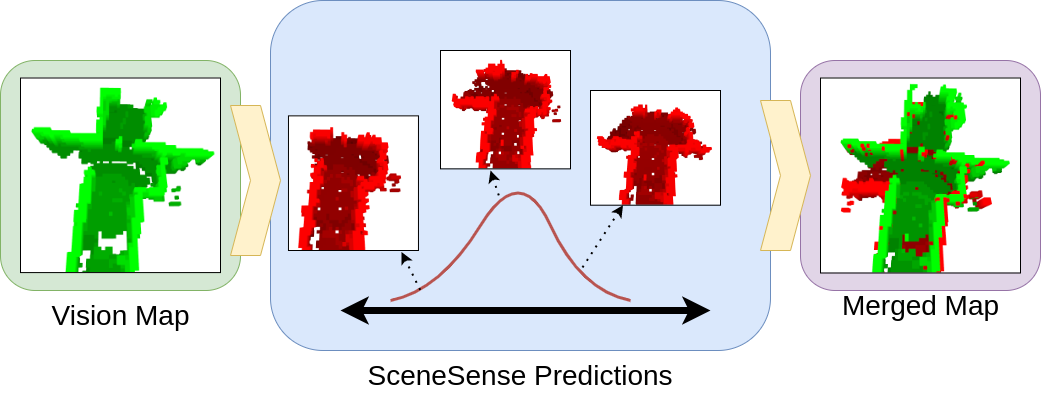}
    \caption{\textbf{Multi-Prediction Occupancy Merging}: SceneSense predicts various occupancy maps based on equivalent input data that form a distribution. This distribution forms a curve where more likely predictions occur more often, and less likely predictions occur infrequently. These predictions are fused into the merged map using Eq. \ref{eq:prob_map_merging}. The resulting merged map naturally filters out the unlikely voxel predictions, forming an extended occupancy map.}
    \label{fig:occ_voting}
\end{figure}
\begin{figure}[t!]
\vspace{5pt}
    \centering
    % First row of subfigures
    \begin{subfigure}[b]{0.15\textwidth}
        \includegraphics[width=\textwidth]{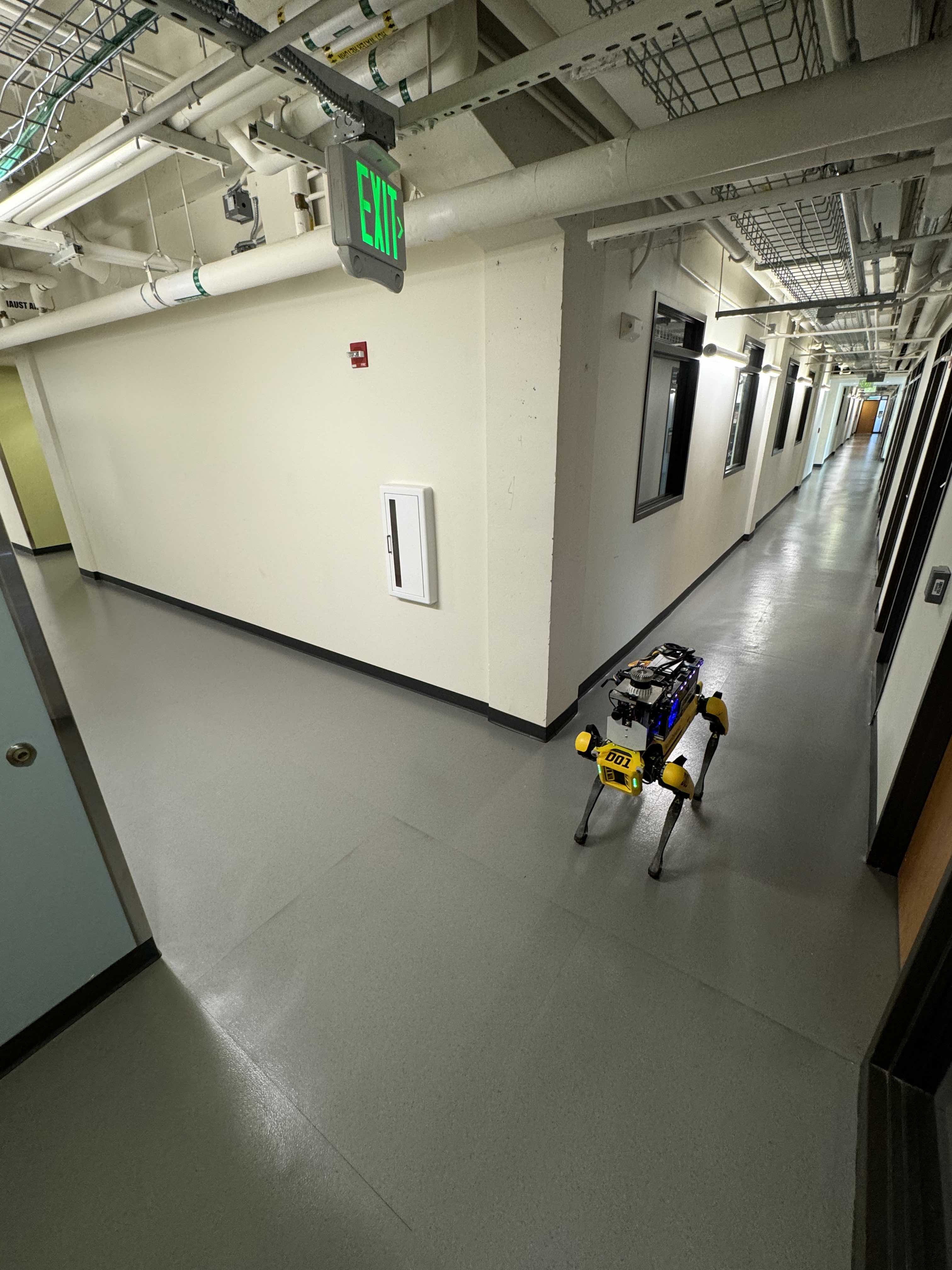}
        \caption{}
        \label{fig:example_preds_a}
    \end{subfigure}
    \hfill
    \begin{subfigure}[b]{0.15\textwidth}
        \includegraphics[width=\textwidth]{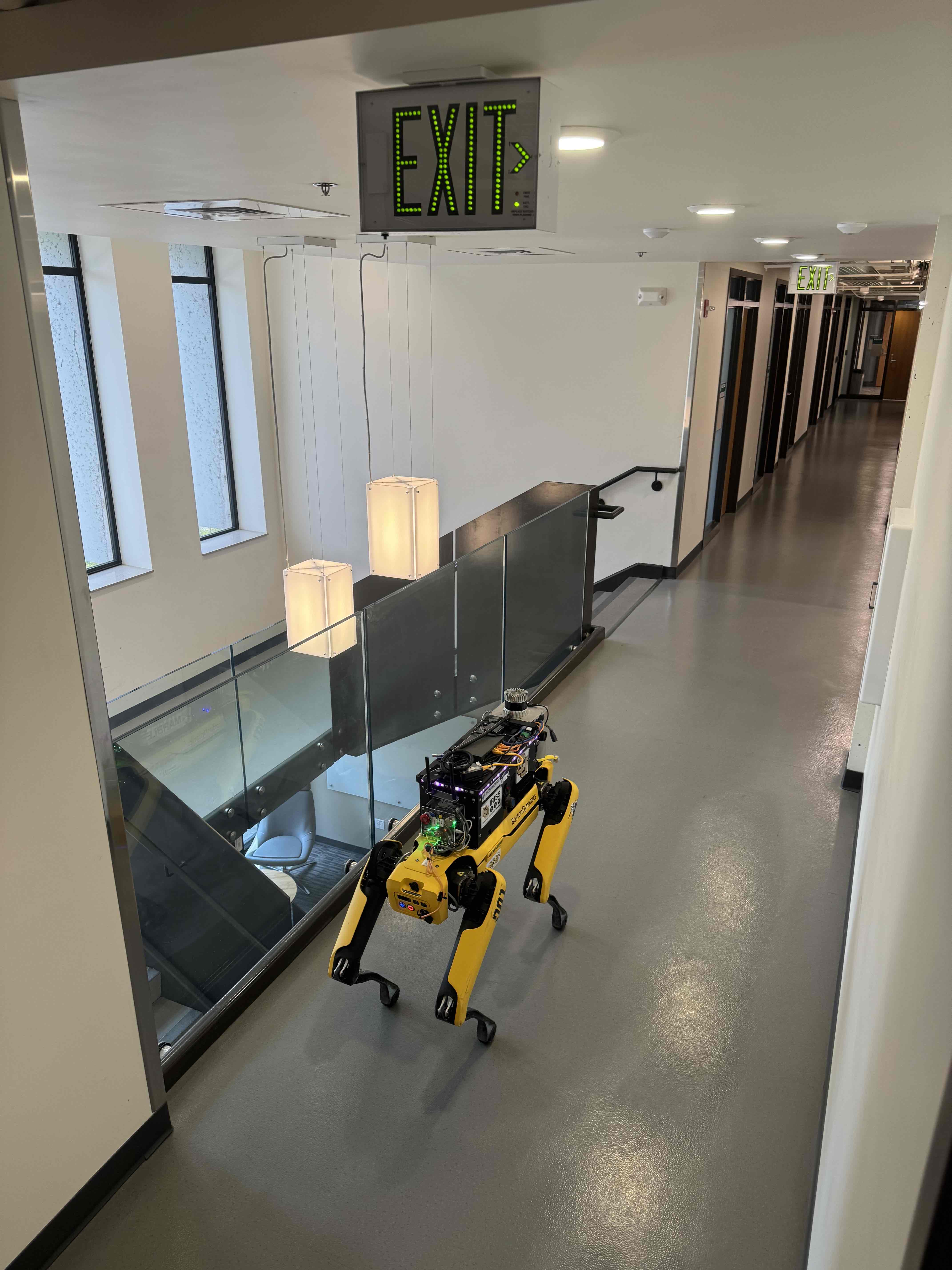}
        \caption{}
        \label{fig:example_preds_b}
    \end{subfigure}
    \hfill
    \begin{subfigure}[b]{0.15\textwidth}
        \includegraphics[width=\textwidth]{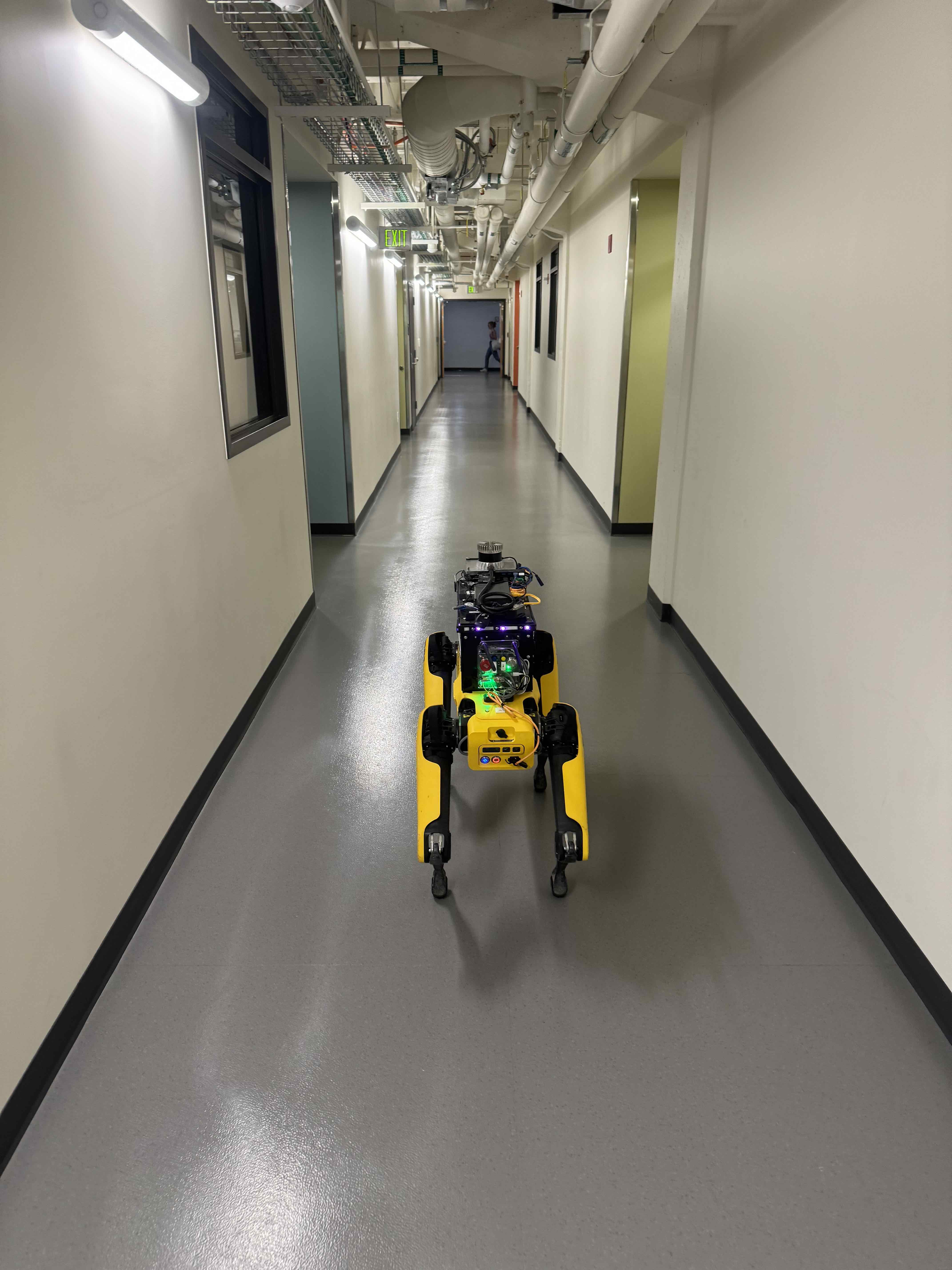}
        \caption{}
        \label{fig:example_preds_c}
    \end{subfigure}

    \vspace{0em} % Space between rows

    % Second row of subfigures
    \begin{subfigure}[b]{0.23\textwidth}
        \includegraphics[width=\textwidth]{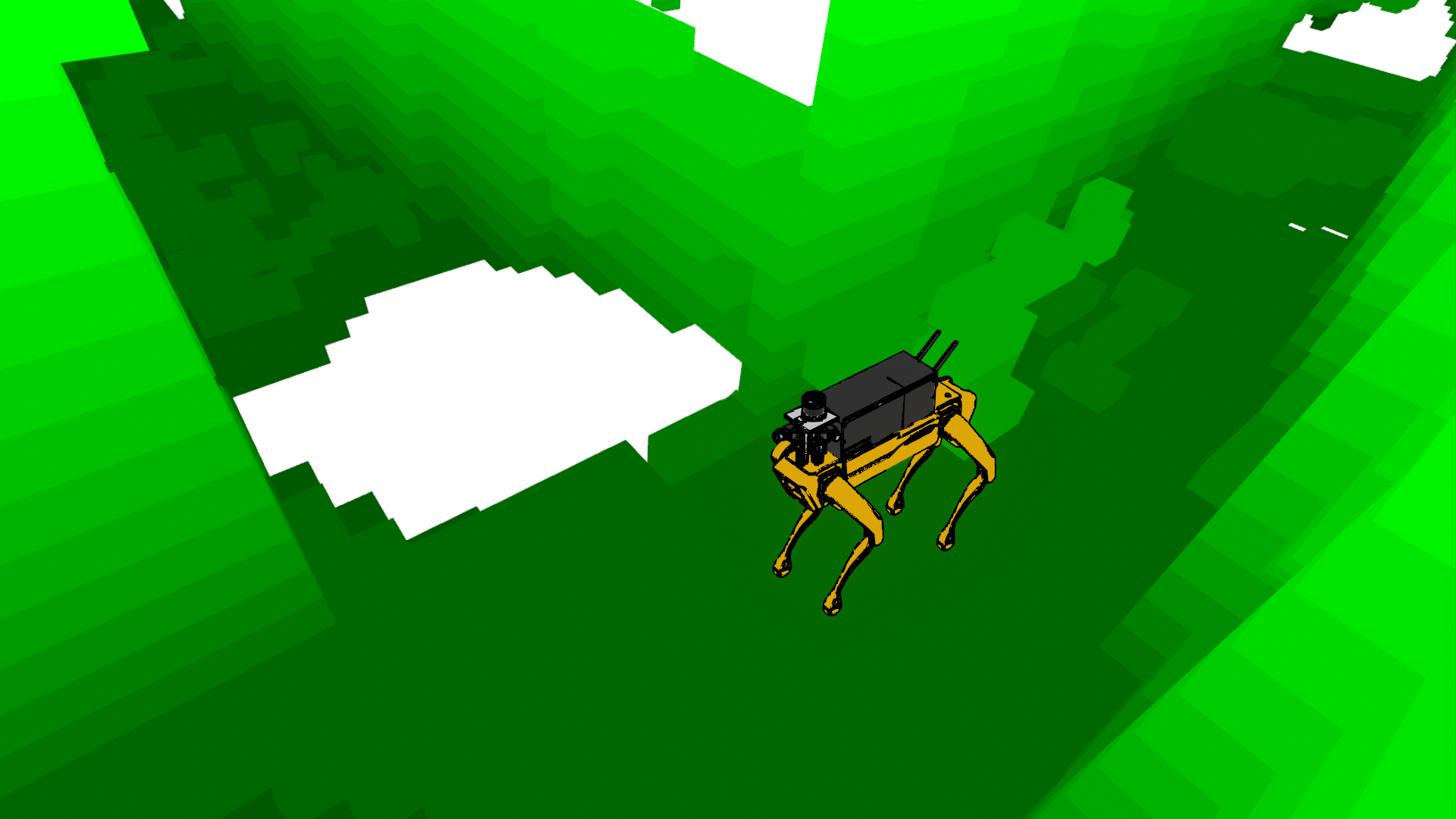}
        % \caption{Subfigure 4}
    \end{subfigure}
    \hfill
    \begin{subfigure}[b]{0.23\textwidth}
        \includegraphics[width=\textwidth]{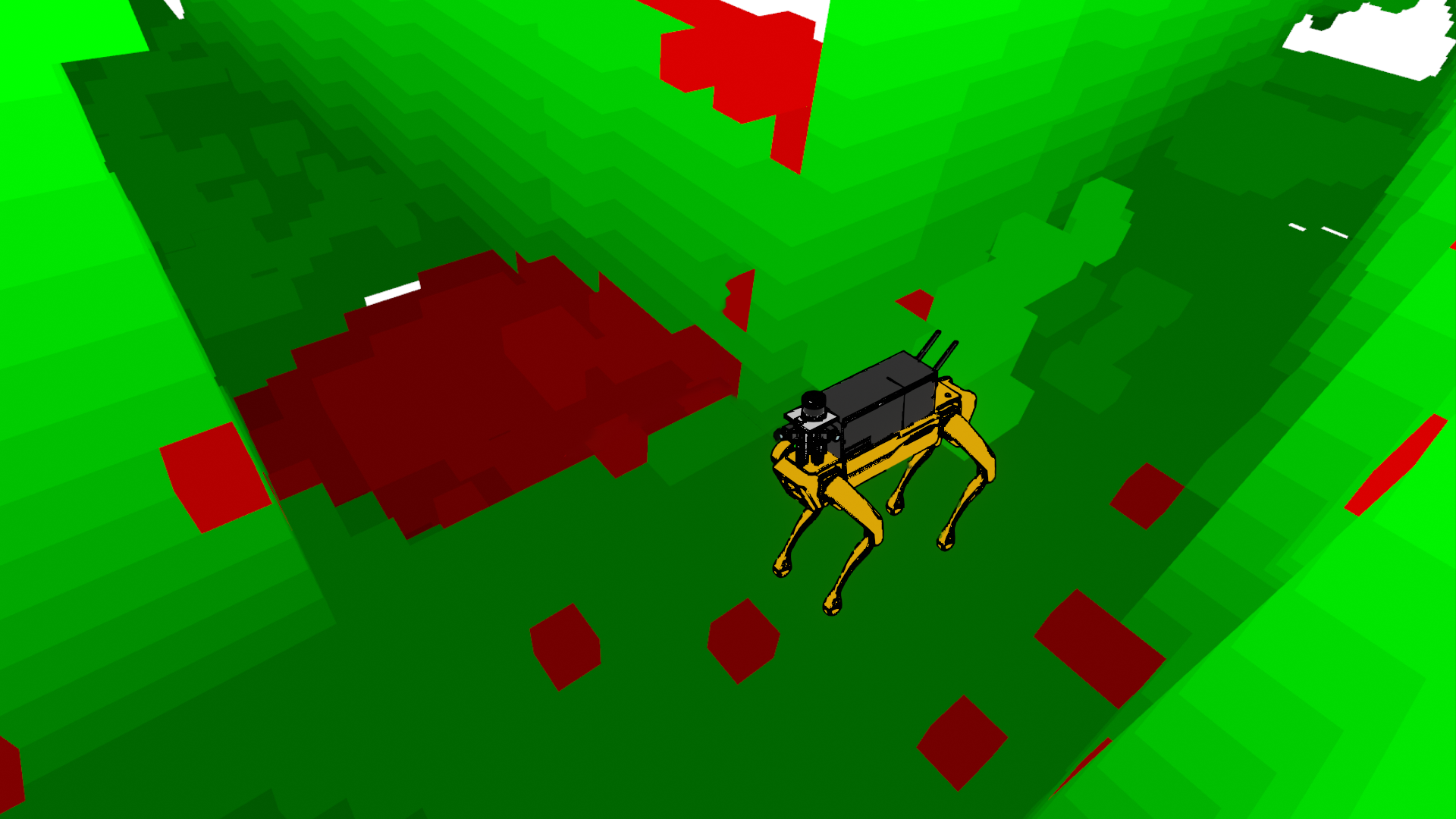}
        % \caption{Subfigure 5}
    \end{subfigure}
    
    % \vspace{-0.8em} % Space between rows

    % Second row of subfigures
    \begin{subfigure}[b]{0.23\textwidth}
        \includegraphics[width=\textwidth]{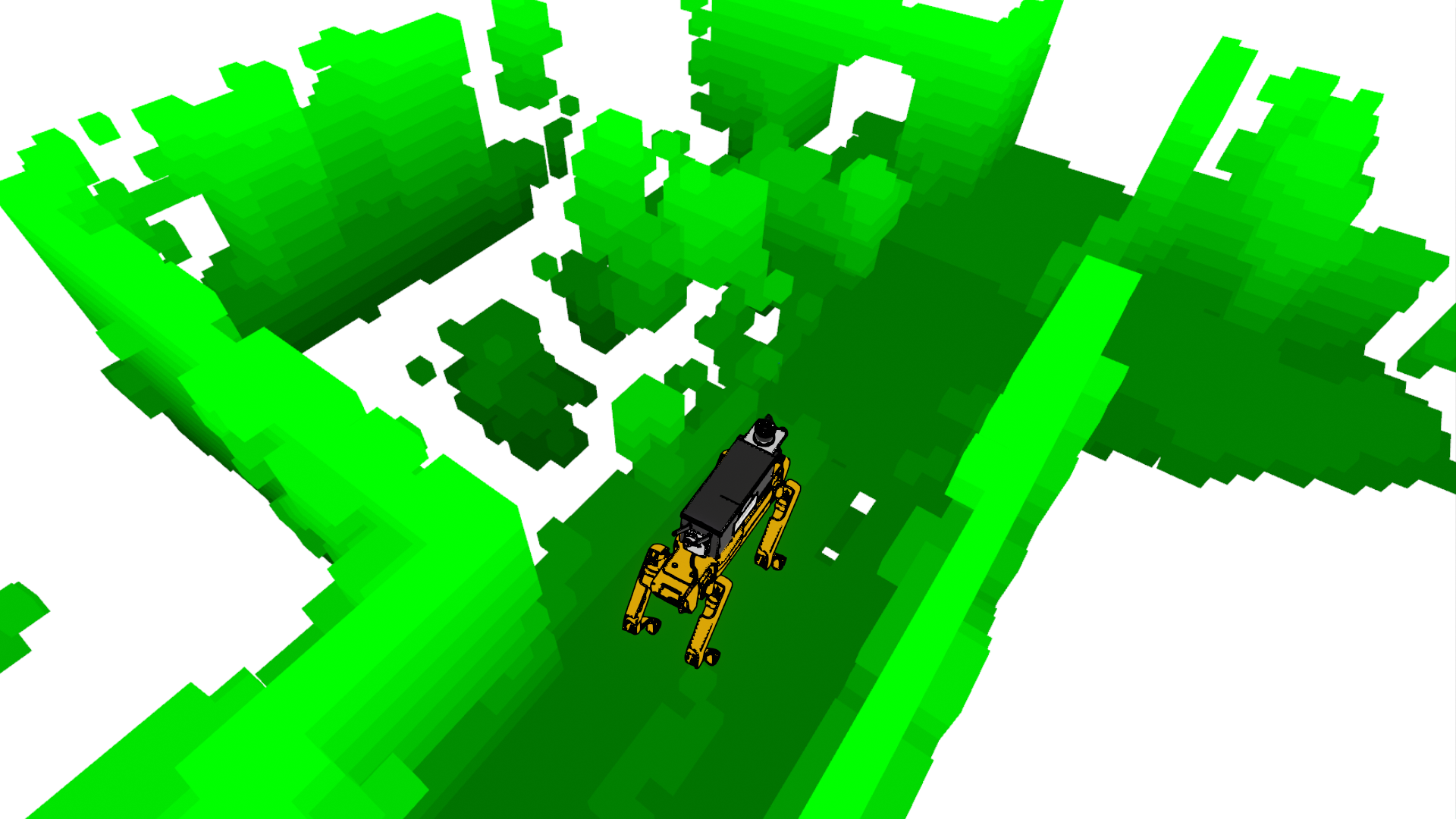}
        % \caption{Subfigure 4}
    \end{subfigure}
    \hfill
    \begin{subfigure}[b]{0.23\textwidth}
        \includegraphics[width=\textwidth]{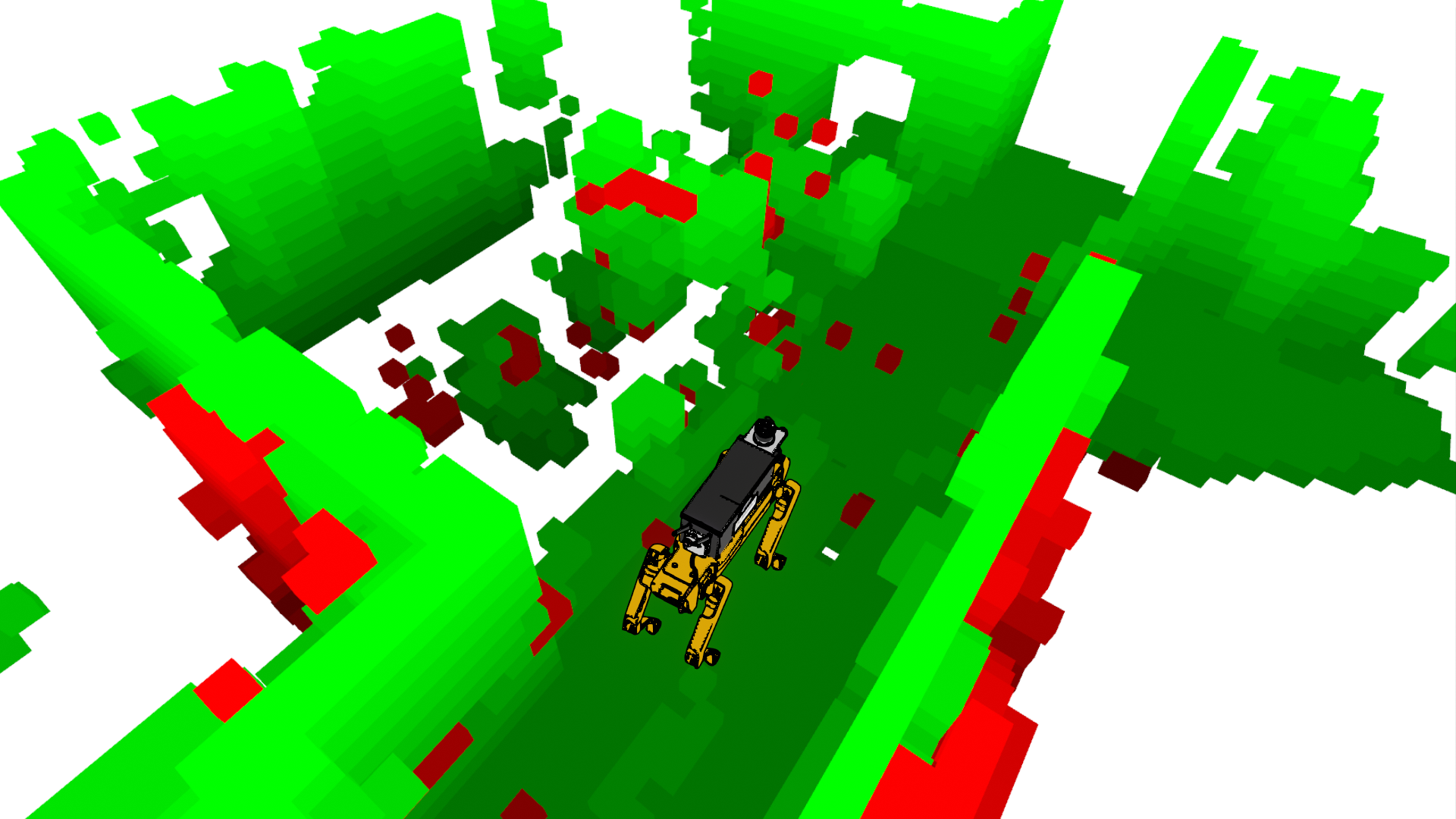}
        % \caption{Subfigure 5}
    \end{subfigure}
    % \vspace{-0.8em} % Space between rows

    % Second row of subfigures
    \begin{subfigure}[b]{0.23\textwidth}
        \includegraphics[width=\textwidth,trim={4cm 0cm 8cm 0cm},clip]{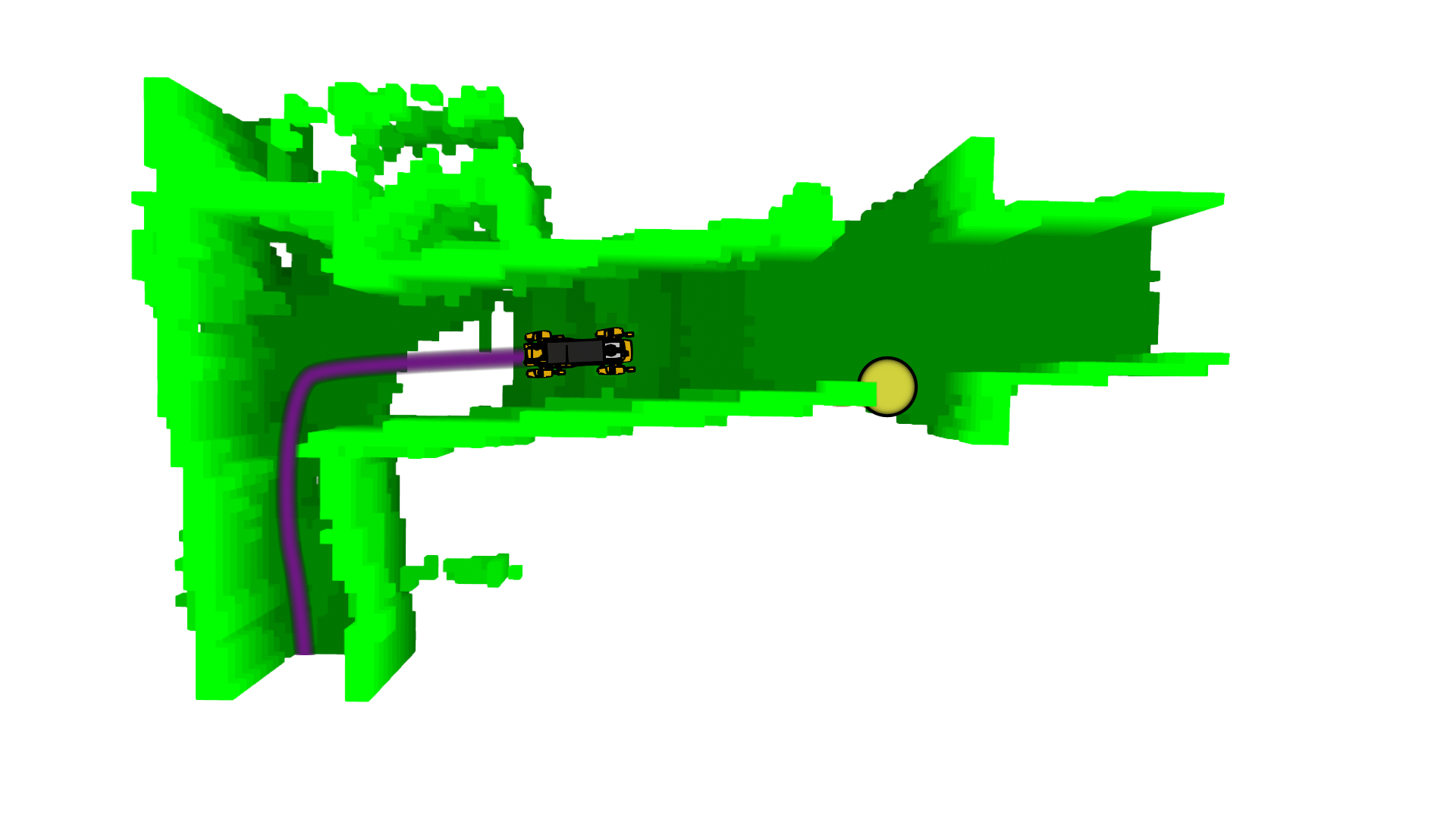}
        % \caption{Subfigure 4}
    \end{subfigure}
    \hfill
    \begin{subfigure}[b]{0.23\textwidth}
        \includegraphics[width=\textwidth,trim={4cm 0cm 8cm 0cm},clip]{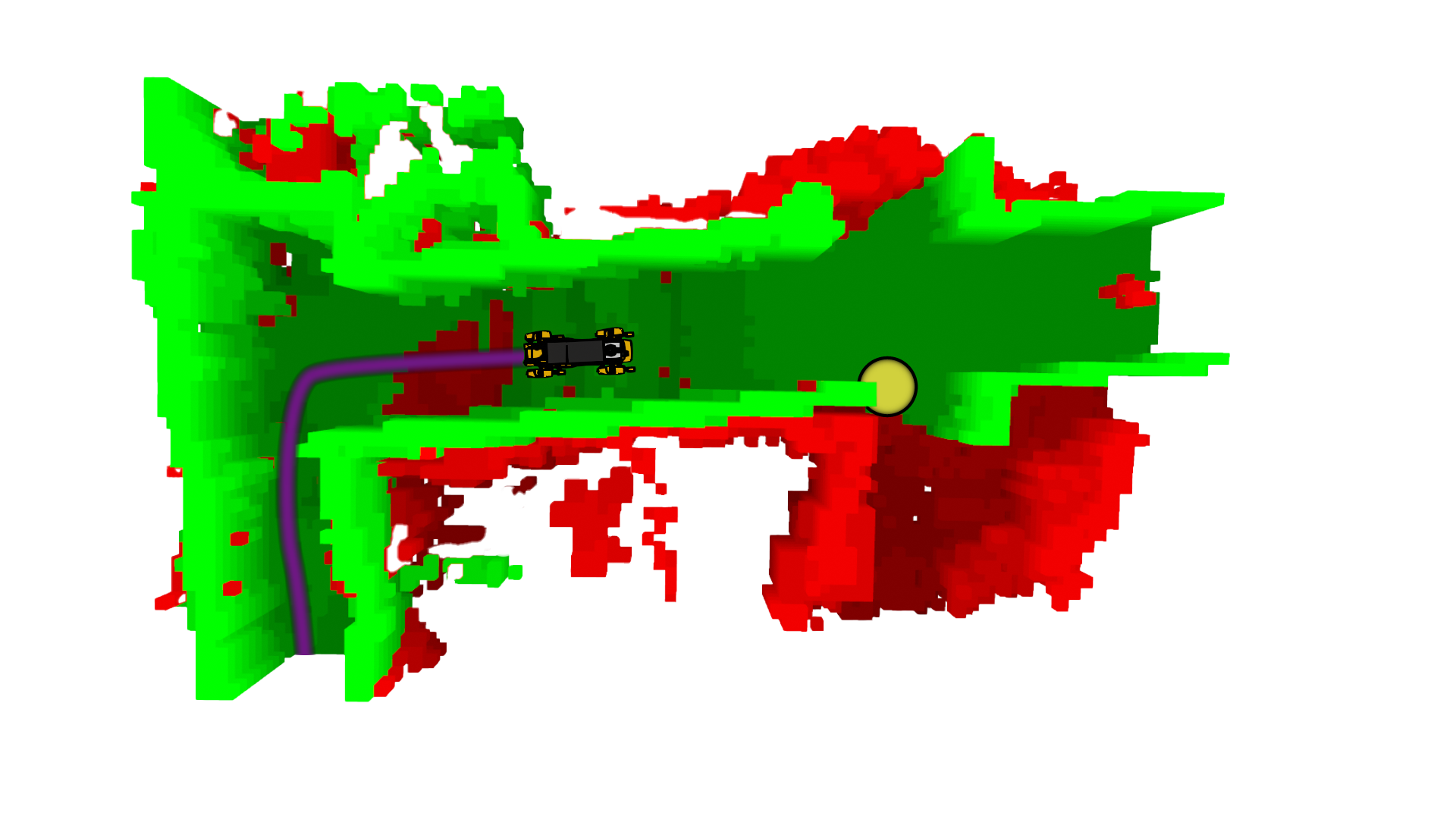}
        % \caption{Subfigure 5}
    \end{subfigure}

    \caption{\textbf{Example Occupancy Predictions}:  Scene images at the top of the figure correspond to the 3 pairs of occupancy maps, where (a) corresponds to the top pair of occupancy maps. The left column of occupancy maps shows the vision only map, while the the right column shows the merged vision and prediction maps. \textbf{(a)} Spot approaches a hallway corner and given the LIDAR mounting position cannot observe the floor after entering the hall junction. SceneSense is able to fill the floor as we well as missing wall information that was not observed. \textbf{(b)} Spot navigates down a hallway and enters an area with a glass railing above the stairs. SceneSense does not fill the open space, where algorithms like hole filling or normal ground expansion may fail. \textbf{(c)}. Spot navigates down a hallway generating predictions along the way. Spot's trajectory is shown in purple, and the identified frontier point is shown in yellow. Beyond providing predictions for the areas that have already been observed, SceneSense generates a frontier prediction at the 4-way intersection. This prediction shows the left side to be a dead-end, while a hallway or entryway is predicted on the right. In reality, these halls are really classrooms, where doors may be open or closed to allow for robot traversal.}
    \label{fig:example_predictions}
\end{figure}
\vspace{5pt}\\
\textbf{Multi-Prediction Occupancy Refinement}. \ \
As shown in Figure \ref{fig:occ_voting}, the occupancy predictions generated by SceneSense can be unique even given the same conditioning information. Similar to image generation tasks it is desirable for SceneSense to generate various realistic results given the same input data \cite{bińkowski2018demystifying, rombach2022high}. Given this behavior, we can collect various predictions from the same location forming a distribution. Then we can aggregate the predictions using the probabilistic update rule defined in Eq. \ref{eq:prob_map_merging} and generate a map constructed from the distribution. The resulting merged map will naturally filter out outlier occupancy predictions and result in only the most probable voxels maintaining occupancy in the final merged map. 
\vspace{5pt}\\
\textbf{Observed vs. Predicted Voxels}. \ \
 SceneSense uses observed data (observed occupied and observed unoccupied) for occupancy inpainting during diffusion. In this paradigm SceneSense will never modify observed cells, and only make occupancy prediction in unobserved space. In Eq. \ref{eq:prob_map_merging} if a voxel has not been observed, SceneSense will generate occupancy predictions and update the voxel given the update rule. If that voxel is later observed, the previous $P(m \mid j_{1:t})$ is used to calculate the probability of occupancy given $P(n \mid z_t)$. In practice it is often the case that the user trusts the LIDAR sensor more than the SceneSense predictions and therefore would configure $P(m \mid d_t)$ $<$ $P(m \mid z_t)$. This means that when a voxel that has been previously populated by SceneSense is directly observed $P(m \mid j_{1:t})$ it will more quickly be updated to reflect the occupancy observed by sensor $z_t$.

Furthermore this approach ensures that SceneSense will never overwrite direct observations. Observed occupancy information (occupied information from LIDAR hits, and unoccupied information calculated from ray casting) is mapped into the diffusion process at every timestep $t$ to perform occupancy inpainting. Therefore any observations, whether those observations be occupied or unoccupied are maintained and guaranteed to persist through the diffusion process. 
\section{EXPERIMENTS AND RESULTS}
\label{sec:experiments_results}
In this section, we provide results and evaluations of the modified SceneSense occupancy prediction framework onboard a real-world system.
\vspace{5pt}\\
\textbf{Training and Implementation}. \ \
To train SceneSense we collected real-world occupancy maps from various buildings. We gathered approximately 1 hour worth of occupancy data, resulting in $11,296$ unique poses with associated complete local occupancy maps. Any areas that were used to train the model are omitted from the results presented here. 

We implement the same denoising network structure presented in \cite{reed2024scenesense}. It is a U-net constructed from the HuggingFace Diffusers library of blocks \cite{von-platen-etal-2022-diffusers} and consists of Resnet \cite{he2015resnet} downsampling/upsampling blocks. The diffusion model is trained using randomly shuffled pairs of ground truth local occupancy maps $x$. We use Chameleon cloud computing resources \cite{keahey2020chameleon} to train our model on one A100 with a batch size of $32$ for $250$ epochs or $88,208$ training steps. We use a cosine learning rate scheduler with a 500 step warm up from $10^{-6}$ to $10^{-4}$. The noise scheduler for diffusion is set to $1,000$ noise steps. 

At inference time we evaluate our dataset using an RTX 4070 TI Super GPU for acceleration. The number of diffusion steps is configured to $30$ steps.
\subsection{Inference time}
 We evaluate the inference time of the unconditional diffusion model against the inference time of the conditional model presented in the original SceneSense paper \cite{reed2024scenesense}. The cross-attention enabling trainable parameters are removed for the unconditional model, but the number of output channels for the constructed U-net are held constant between both models. As the ablation results of the original paper show minor, or no performance gain between the conditional and unconditional model in this configuration we do not evaluate the results of the model predictions in these experiments. 
\begin{table}[h]
\begin{center}
\caption{Inference time and model size results. ``Full inference'' and ``end-to-end'' evaluations are computed using 30 diffusion steps.}
\label{table:inference_time}
\begin{tabular}{p{2cm} | p{2.5cm}  p{2cm} } 
\hline
 & Cond. Model \cite{reed2024scenesense} & Uncond. Model \\
\hline
Trainable Params & 141,125,261 & \textbf{101,144,845}\\
Diffusion Step (s) & 0.03707 & \textbf{0.0147} \\
Full Inference (s) & 1.11 & \textbf{0.4437} \\
Backbone (s) & .55099 & N.A. \\
End-to-end (s) & 1.66 & \textbf{0.4437} \\

\end{tabular}
\end{center}
\end{table}
% pointnet++ eval time: .55099 
\vspace{5pt}\\
\textbf{Discussion}. \ \  As shown in Table \ref{table:inference_time}, 
%and as hypothesized by the SceneSense authors \cite{reed2024scenesense} 
removing the conditioning from the diffusion model reduces the computation requirements substantially. The unconditioned model reduces the number of trainable parameters by $28\%$, the model inference time by $60\%$ and the end to end computation time by $73\%$. These improvements enable SceneSense to operate in real-time more effectively, allowing for more flexible implementations for onboard robotic applications.
\subsection{SceneSense Generative Occupancy Evaluation}
For the following experiments we evaluated the occupancy generation capabilities of SceneSense onboard a real world robot in 2 unique test environments. In particular, we examine the fidelity of predictions around the robot with predictions at the frontiers of the map, ablating the map update methods and the running sensor only map.
\vspace{5pt}\\
\textbf{Experimental Setup}. \ \ SceneSense predictions are evaluated in 2 environments. Environment 1 was a long hallway with cutouts for classrooms and 1 right turn. Select frames shown in figure \ref{fig:example_preds_a} and \ref{fig:example_preds_c} are from Environment 1. Environment 2 consists of similar carpeted area with 4 hallways and 4 turns, forming a square shape. We evaluate the occupancy prediction framework using the following test configurations. 
\begin{enumerate}
    \item \textbf{Baseline or SceneSense}: A comparison between octomap sensor only local occupancy (BL) with the SceneSense occupancy prediction included (SS). 
    \item \textbf{Robot-centric or Frontier-centric}: Robot-centric diffusion (RC) predicts occupancy at a radius of $3.3m$ about the robot while frontier-centric diffusion (FC) predicts occupancy at a radius of $3.3m$ at an identified location in the map, which has a maximum range of 7$m$ from the robot. 
    \item \textbf{One Shot Map Merging or Probabilistic Map Merging}: One shot map merging (OSMM) simply takes the current local occupancy map and a SceneSense occupancy prediction and populates the predicted occupancy information in the running map. Probabilistic map merging (PMM) keeps a running local merged occupancy map that uses update equation \ref{eq:prob_map_merging} to update the occupancy map for every new occupancy prediction. In practice, each pose will receive 3-5 SceneSense predictions to merge into the running map.
\end{enumerate}
\vspace{5pt}
\textbf{Occupancy Prediction Metrics}. \ \ 
Following similar generative scene synthesis approaches \cite{tang2023diffuscene, wang2021sceneformer}  we employ the Fr\'echet inception distance (FID) \cite{NIPS2017_8a1d6947} and the Kernel inception distance \cite{bińkowski2018demystifying} (KID $\times 1000$) to evaluate the generated local occupancy grids using the clean-fid library \cite{parmar2021cleanfid}. Generating good metrics to evaluate generative frameworks is a difficult task \cite{naeem2020reliable}. FID and KID have become the standard metric for many generative methods due to their ability to score both accuracy of predicted results, and diversity or coverage of the results when compared to a set of ground truth data. While these metrics are fairly new to robotics, which traditionally evaluates occupancy data with metrics like accuracy, precision and IoU, these metrics have been shown to be an effective measure for evaluating predicted scenes \cite{reed2024scenesense, tang2023diffuscene}.
\begin{table}[h]
\begin{center}
\caption{Results comparing running occupancy (BL) to occupancy enhanced with SceneSense prediction (SS). Evaluations of each method are provided as robot-centric generations (RC) and frontier-centric generations (FC).}
\label{table:sim_results}
\begin{tabular}{l | c c | c c } 
\hline
 & \multicolumn{2}{c}{Env. 1} & \multicolumn{2}{c}{Env. 2}\\
Method & FID  $\downarrow$ & KID$\times$1000 $\downarrow$ & FID  $\downarrow$ & KID$\times$1000  $\downarrow$ \\
\hline
% 3D Sketch \cite{chen20203dsketch} & & & &\\
BL-RC & 36.0 & 16.4 & 30.3 &  16.3\\ 
SS-RC-OSMM &\textbf{26.3}  & \textbf{7.7} & \textbf{20.8} & 10.1 \\
SS-RC-PMM & 29.2 &  10.4& 21.0  & \textbf{9.1}\\
\hline 
BL-FC & 116.9 & 81.6 & 150.6 & 118.8 \\ 
SS-FC-OSMM & 104.2 & 66.3 & 133.4  & 104.4 \\
SS-FC-PMM & \textbf{30.1} & \textbf{10.3} & \textbf{34.5}  & \textbf{9.0}\\
\hline
\end{tabular}
\end{center}
\end{table}

\textbf{Results Discussion}. \ \ The results in Table \ref{table:sim_results} show that RC predictions are quite similar between OSMM and PMM approaches, reducing the FID of the environments by an average of $28.5\%$ and $25\%$ for OSMM and PMM  respectively. These results are similar to the simulation-based results presented in \cite{reed2024scenesense}. However, the FC results show a much greater improvement for PMM, with an average FID reduction of $75\%$, compared to OSMM, which only achieves an average FID reduction of $11\%$.

Interestingly, The KID values are nearly identical between between SS-RC-PMM and SS-FC-PMM, indicating that the model occupancy predictions at range are as reasonable as the predictions made around the robot, even though there is less information for the predictions at range. KID is known to be less sensitive to outliers when compared to FID  \cite{bińkowski2018demystifying}. It is likely that the unreasonable predictions that can occur when performing FC predictions are better filtered out by the KID metric, resulting in similar scores.

The large discrepancy between OSMM and PMM results when evaluated at frontiers is due to the sparsity of the occupancy map at the frontier. We predict that as the number of unknown voxels grow, so too does the distribution of predicted scenes. Intuitively, if there are no observed voxels to guide the prediction SceneSense will predict a wide variety of possible occupancy maps. On the other end, if all voxels in the target space are observed, the same occupancy map will be generated every time.

We can analyze the number of available voxels for occupancy prediction as the number of unknown voxels in the target area $x_{rm}$ as a percentage of the total observed voxels in $x_{gt}$.  Using the results from environment 2 to evaluate this prediction, we calculate that on average $59.18\%$ of target area voxels are unknown when performing RC occupancy prediction. However, when performing FC prediction this number jumps to $70.79\%$. This result supports the intuitive statement that there are more available (unknown) voxels for prediction around the frontiers of the map than around the robot. 
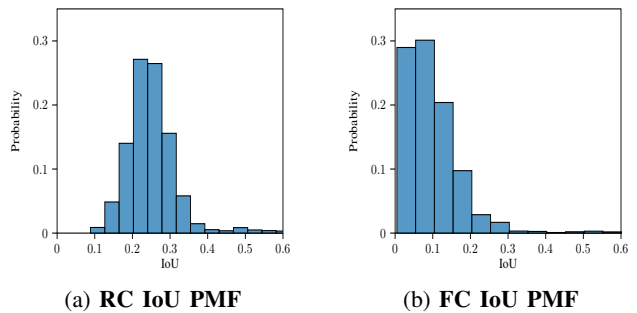
\begin{figure}[t]
\vspace{6pt}
\centering
    \hfill\subfloat[][\textbf{RC IoU PMF}]{\resizebox{110pt}{100pt}{% This file was created with tikzplotlib v0.10.1.
\begin{tikzpicture}

\definecolor{darkgray176}{RGB}{176,176,176}
\definecolor{steelblue31119180}{RGB}{31,119,180}

\begin{axis}[
tick align=outside,
tick pos=left,
x grid style={darkgray176},
xlabel={IoU},
xmin=0, xmax=0.6,
xtick style={color=black},
y grid style={darkgray176},
ylabel={Probability},
ymin=0, ymax=0.35,
ytick style={color=black}
]
\draw[draw=black,fill=steelblue31119180,fill opacity=0.75] (axis cs:0.0887610437641257,0) rectangle (axis cs:0.126723146674795,0.00879120879120879);
\draw[draw=black,fill=steelblue31119180,fill opacity=0.75] (axis cs:0.126723146674795,0) rectangle (axis cs:0.164685249585464,0.0485958485958486);
\draw[draw=black,fill=steelblue31119180,fill opacity=0.75] (axis cs:0.164685249585464,0) rectangle (axis cs:0.202647352496133,0.14017094017094);
\draw[draw=black,fill=steelblue31119180,fill opacity=0.75] (axis cs:0.202647352496133,0) rectangle (axis cs:0.240609455406803,0.271306471306471);
\draw[draw=black,fill=steelblue31119180,fill opacity=0.75] (axis cs:0.240609455406803,0) rectangle (axis cs:0.278571558317472,0.264713064713065);
\draw[draw=black,fill=steelblue31119180,fill opacity=0.75] (axis cs:0.278571558317472,0) rectangle (axis cs:0.316533661228141,0.155799755799756);
\draw[draw=black,fill=steelblue31119180,fill opacity=0.75] (axis cs:0.316533661228141,0) rectangle (axis cs:0.35449576413881,0.0581196581196581);
\draw[draw=black,fill=steelblue31119180,fill opacity=0.75] (axis cs:0.35449576413881,0) rectangle (axis cs:0.39245786704948,0.0146520146520147);
\draw[draw=black,fill=steelblue31119180,fill opacity=0.75] (axis cs:0.39245786704948,0) rectangle (axis cs:0.430419969960149,0.00537240537240537);
\draw[draw=black,fill=steelblue31119180,fill opacity=0.75] (axis cs:0.430419969960149,0) rectangle (axis cs:0.468382072870818,0.00366300366300366);
\draw[draw=black,fill=steelblue31119180,fill opacity=0.75] (axis cs:0.468382072870818,0) rectangle (axis cs:0.506344175781487,0.00854700854700855);
\draw[draw=black,fill=steelblue31119180,fill opacity=0.75] (axis cs:0.506344175781487,0) rectangle (axis cs:0.544306278692157,0.00488400488400488);
\draw[draw=black,fill=steelblue31119180,fill opacity=0.75] (axis cs:0.544306278692157,0) rectangle (axis cs:0.582268381602826,0.00390720390720391);
\draw[draw=black,fill=steelblue31119180,fill opacity=0.75] (axis cs:0.582268381602826,0) rectangle (axis cs:0.620230484513495,0.00317460317460317);
\draw[draw=black,fill=steelblue31119180,fill opacity=0.75] (axis cs:0.620230484513495,0) rectangle (axis cs:0.658192587424164,0.00268620268620269);
\draw[draw=black,fill=steelblue31119180,fill opacity=0.75] (axis cs:0.658192587424165,0) rectangle (axis cs:0.696154690334834,0.00146520146520147);
\draw[draw=black,fill=steelblue31119180,fill opacity=0.75] (axis cs:0.696154690334834,0) rectangle (axis cs:0.734116793245503,0.00244200244200244);
\draw[draw=black,fill=steelblue31119180,fill opacity=0.75] (axis cs:0.734116793245503,0) rectangle (axis cs:0.772078896156172,0.000732600732600733);
\draw[draw=black,fill=steelblue31119180,fill opacity=0.75] (axis cs:0.772078896156172,0) rectangle (axis cs:0.810040999066841,0);
\draw[draw=black,fill=steelblue31119180,fill opacity=0.75] (axis cs:0.810040999066841,0) rectangle (axis cs:0.848003101977511,0.000976800976800977);
\end{axis}

\end{tikzpicture}}}\hfill
    \quad
    \hfill\subfloat[][\textbf{FC IoU PMF}]{\resizebox{110pt}{100pt}{% This file was created with tikzplotlib v0.10.1.
\begin{tikzpicture}

\definecolor{darkgray176}{RGB}{176,176,176}
\definecolor{steelblue31119180}{RGB}{31,119,180}

\begin{axis}[
tick align=outside,
tick pos=left,
x grid style={darkgray176},
xlabel={IoU},
xmin=0, xmax=0.6,
xtick style={color=black},
y grid style={darkgray176},
ylabel={Probability},
ymin=0, ymax=0.35,
ytick style={color=black}
]
\draw[draw=black,fill=steelblue31119180,fill opacity=0.75] (axis cs:0.00477232054086299,0) rectangle (axis cs:0.0545337045138199,0.289768483943241);
\draw[draw=black,fill=steelblue31119180,fill opacity=0.75] (axis cs:0.0545337045138199,0) rectangle (axis cs:0.104295088486777,0.301157580283794);
\draw[draw=black,fill=steelblue31119180,fill opacity=0.75] (axis cs:0.104295088486777,0) rectangle (axis cs:0.154056472459734,0.203883495145631);
\draw[draw=black,fill=steelblue31119180,fill opacity=0.75] (axis cs:0.154056472459734,0) rectangle (axis cs:0.20381785643269,0.0974607916355489);
\draw[draw=black,fill=steelblue31119180,fill opacity=0.75] (axis cs:0.20381785643269,0) rectangle (axis cs:0.253579240405647,0.0287528005974608);
\draw[draw=black,fill=steelblue31119180,fill opacity=0.75] (axis cs:0.253579240405647,0) rectangle (axis cs:0.303340624378604,0.0168035847647498);
\draw[draw=black,fill=steelblue31119180,fill opacity=0.75] (axis cs:0.303340624378604,0) rectangle (axis cs:0.353102008351561,0.00317401045556385);
\draw[draw=black,fill=steelblue31119180,fill opacity=0.75] (axis cs:0.353102008351561,0) rectangle (axis cs:0.402863392324518,0.00261389096340553);
\draw[draw=black,fill=steelblue31119180,fill opacity=0.75] (axis cs:0.402863392324518,0) rectangle (axis cs:0.452624776297475,0.000933532486930545);
\draw[draw=black,fill=steelblue31119180,fill opacity=0.75] (axis cs:0.452624776297475,0) rectangle (axis cs:0.502386160270432,0.0020537714712472);
\draw[draw=black,fill=steelblue31119180,fill opacity=0.75] (axis cs:0.502386160270432,0) rectangle (axis cs:0.552147544243388,0.00298730395817774);
\draw[draw=black,fill=steelblue31119180,fill opacity=0.75] (axis cs:0.552147544243388,0) rectangle (axis cs:0.601908928216345,0.00186706497386109);
\draw[draw=black,fill=steelblue31119180,fill opacity=0.75] (axis cs:0.601908928216345,0) rectangle (axis cs:0.651670312189302,0.000373412994772218);
\draw[draw=black,fill=steelblue31119180,fill opacity=0.75] (axis cs:0.651670312189302,0) rectangle (axis cs:0.701431696162259,0.00168035847647498);
\draw[draw=black,fill=steelblue31119180,fill opacity=0.75] (axis cs:0.701431696162259,0) rectangle (axis cs:0.751193080135216,0.00149365197908887);
\draw[draw=black,fill=steelblue31119180,fill opacity=0.75] (axis cs:0.751193080135216,0) rectangle (axis cs:0.800954464108173,0.00373412994772218);
\draw[draw=black,fill=steelblue31119180,fill opacity=0.75] (axis cs:0.800954464108173,0) rectangle (axis cs:0.85071584808113,0.00392083644510829);
\draw[draw=black,fill=steelblue31119180,fill opacity=0.75] (axis cs:0.850715848081129,0) rectangle (axis cs:0.900477232054086,0.00896191187453323);
\draw[draw=black,fill=steelblue31119180,fill opacity=0.75] (axis cs:0.900477232054086,0) rectangle (axis cs:0.950238616027043,0.0125093353248693);
\draw[draw=black,fill=steelblue31119180,fill opacity=0.75] (axis cs:0.950238616027043,0) rectangle (axis cs:1,0.0158700522778193);
\end{axis}

\end{tikzpicture}}}\hfill
    \quad
    \caption{\textbf{Env 2 IoU Probability Mass Function (PMF)}. (a) IoU histogram of RC SceneSense predictions. (b) IoU histogram of FC SceneSense predictions. The IoU distributions show that RC occupancy predictions are more likely to be similar than FC predictions.}
    \label{fig:pred_PMF}
    % \vspace{-20pt}
\end{figure}
To confirm that the increase in unknown voxels widens the distribution of occupancy prediction we generate a distribution of results by calculating the IoU of each prediction against all other predictions made during the run. The results of this approach as provided in Figure \ref{fig:pred_PMF} show that RC predictions are more likely to be similar, while FC predictions are more likely to be dissimilar with very little overlap. When predictions are all similar, PMM becomes less important for accurate predictions, since OSMM would result in a similar map each time. However PMM is needed at range to achieve reasonable results since it can negotiate the wider distribution of possible occupancy predictions. 
% \vspace{5pt}\\
% \textbf{Implementation Discussion}. \ \ 
% The primary challenge associated with adapting SceneSense from simulation to real-world was the noise introduced in the lidar scans, and in particular how errant scans impacted the Octomap ray-casting algorithm. With the current implementation of \emph{occupancy inpainting}, SceneSense is designed to treat observed voxels as ground truth occupancy, propagating any vision errors to the occupancy prediction. Not only does this error propagation keep SceneSense from predicting occupancy in desirable locations, if enough errors accumulate it can result in inpainting that moves the occupancy prediction out of the training distribution, resulting in worse predictions. These vision errors become more common as the lidar beam length increases. We found through testing that 7$m$ was the maximum beam length where we could consistently generate reasonable maps. 
% 26% decrese in FID for OSMM
% 18.8% for PMM
% 31% for both OSMM and PMM  in ENV 2
% 10.8% FC-OSMM ENV 1
% 74% FC-PMM ENV 1
% 11% FC-OSMM ENV 2
% 77% FC-PMM ENV 2

\section{CONCLUSIONS}
In this work we present key architectural changes to the SceneSense \cite{reed2024scenesense} occupancy prediction model to enable real time occupancy inference at any location of interest in the map. Further we present our integration of SceneSense to a real robotic system, a method for probabilistic merging occupancy predictions into a running occupancy map, as well as evaluations of these occupancy predictions.  Future work will explore how these predictions can be utilized to improve existing planning and exploration architectures.

%%%%%%%%%%%%%%%%%%%%%%%%%%%%%%%%%%%%%%%%%%%%%%%%%%%%%%%%%%%%%%%%%%%%%%%%%%%%%%%%

%%%%%%%%%%%%%%%%%%%%%%%%%%%%%%%%%%%%%%%%%%%%%%%%%%%%%%%%%%%%%%%%%%%%%%%%%%%%%%%%



% \section*{ACKNOWLEDGMENT}

% The preferred spelling of the word ÒacknowledgmentÓ in America is without an ÒeÓ after the ÒgÓ. Avoid the stilted expression, ÒOne of us (R. B. G.) thanks . . .Ó  Instead, try ÒR. B. G. thanksÓ. Put sponsor acknowledgments in the unnumbered footnote on the first page.

%%%%%%%%%%%%%%%%%%%%%%%%%%%%%%%%%%%%%%%%%%%%%%%%%%%%%%%%%%%%%%%%%%%%%%%%%%%%%%%%
\printbibliography

\addtolength{\textheight}{-12cm}   % This command serves to balance the column lengths

\end{document}